\documentclass[journal]{IEEEtran}
\usepackage{graphicx}
\usepackage{wrapfig,lipsum,cleveref}
\usepackage{amssymb,amsfonts,amsmath}
\usepackage{multirow}

\begin{document}

\title{Deep Recurrent Convolutional Neural Network: Improving Performance For Speech Recognition}

\author{Zewang~Zhang,~\IEEEmembership{Student Member,~IEEE,}
        Zheng~Sun,~\IEEEmembership{Student Member,~IEEE,}
        Jiaqi~Liu,~\IEEEmembership{Student Member,~IEEE,}
        Jingwen~Chen,~\IEEEmembership{Student Member,~IEEE,}
        Zhao~Huo,~\IEEEmembership{Member,~IEEE,}
        and~Xiao~Zhang,~\IEEEmembership{Member,~IEEE}% <-this % stops a space
\thanks{Z. Zhang, Z. Sun, J. Liu, J. Chen and X. Zhang are with the Department
of Physics, Sun Yat--Sen University, Guangzhou, 510275 P.R. China
 (e-mail: zhangzw3@mail2.sysu.edu.cn; sunzh6@mail2.sysu.edu.cn;
 liujq33@mail2.sysu.edu.cn; chenjw93@mail2.sysu.edu.cn; 
zhangxiao@mail.sysu.edu.cn).}% <-this % stops a space
\thanks{Z.~Huo is with China University of Political Science and Law (e-mail: huozhao@cupl.edu.cn).}% <-this % stops a space
\thanks{C. H. Lee is with the Institute of High Performance Computing (e-mail: calvin-lee@ihpc.a-star.edu.sg).}
\thanks{Manuscript received MM DD, YYYY; revised MM DD, YYYY.}}

%\markboth{IEEE TRANSACTIONS ON NEURAL AND LEARNING SYSTEM}%

\maketitle

\begin{abstract}
A deep learning approach has been widely applied in sequence modeling problems. In terms of automatic speech recognition (ASR), its performance has significantly been improved by increasing large speech corpus and deeper neural network. Especially, recurrent neural network and deep convolutional neural network have been applied in ASR successfully. Given the arising problem of training speed, we build a novel deep recurrent convolutional network for acoustic modeling and then apply deep residual learning to it. Our experiments show that it has not only faster convergence speed but better recognition accuracy over traditional deep convolutional recurrent network. In the experiments, we compare the convergence speed of our novel deep recurrent convolutional networks and traditional deep convolutional recurrent networks. With faster convergence speed, our novel deep recurrent convolutional networks can reach the comparable performance. We further show that applying deep residual learning can boost the convergence speed of our novel deep recurret convolutional networks. Finally, we evaluate all our experimental networks by phoneme error rate (PER) with our proposed bidirectional statistical n-gram language model. Our evaluation results show that our newly proposed deep recurrent convolutional network applied with deep residual learning can reach the best PER of 17.33\% with the fastest convergence speed on TIMIT database. The outstanding performance of our novel deep recurrent convolutional neural network with deep residual learning indicates that it can be potentially adopted in other sequential problems. 
\end{abstract}

\begin{IEEEkeywords}
Convergence speed, deep recurrent convolutional neural network, deep residual learning, acoustic modeling 
\end{IEEEkeywords}

\IEEEpeerreviewmaketitle

\section{Introduction}

\IEEEPARstart{S}{equence} modeling is an important part of artificial intelligence, and an efficient sequential model can help machine learn how to think as human intelligence and how to interact with the world by sequential actions and logical thinking skills. A major problem faced by sequence modeling is that the deep learning model suffers too much training time, and new powerful sequential models should be invented to meet the demand of both robustness and efficiency.     

Automatic speech recognition(ASR) is designed to transcript human speech into spoken phonemes. ASR has been investigated for several decades. Traditionally, a statistical model of maximum likelihood decoding and maximum mutual information estimation are used for speech recognition \cite{hmm1,hmm2}, the use of gaussian mixture model (GMM) combined with hidden markov model (HMM) for speech recognition had also become predominant several years ago \cite{0hmm3,hmm3,hmm4}. With the spring-up of deep learning, deep neural network (DNN) with HMM states has been shown to outperform the traditional method of GMM-HMM \cite{dnn0,dnn1,dnn2,dnn3,dnn4,dnn5}, thus many new training tricks have been proposed to improve the performance of DNNs for acoustic modeling, such as powerful non-linear activation functions, layer-wise mini-batch training, batch normalization, dropout and fast gradient descent method \cite{other1,other3,other5,other12,other13}.

DNN is very good at exploiting non-linear feature representation, but it lacks internal time dependency and sequential modeling ability, since human speech is a sequential problem with dynamic features to handle. Recurrent neural network (RNN) is a powerful tool for sequential modeling owing to its recurrent hidden states between adjacent time-steps, and DNNs are gradually replaced by RNNs which have been successfully applied in ASR in the last several years. Meanwhile, Deep Long Short-term Memory RNNs and deep bidirectional RNNs are proposed to exploit long time memory in ASR \cite{rnn1,rnn2,rnn4,rnn8,rnn9,rnn10}. Besides, sequence training of RNNs with connectionist temporal classification (CTC) has shown great performance in end-to-end ASR \cite{rnn11,rnn12,rnn13}. Traditional frame-wise cross entropy training needs pre-segmented data by hand, but CTC is an end-to-end training method for RNNs which decodes the output probability distribution into phoneme sequences without requiring pre-segmented training data. RNN has been widely used in ASR, but RNN can't depict very long time dependency because of its vanishing gradient problem, and deeper RNN seems to have little improvement when the number of layers reaches a limit. Although LSTM improves the performance of RNN, an disadvantage of LSTM is that it requires too much computation and stores multiple gating neural responses at each time-step, which will become a computational bottleneck.

Very recently, some other novel neural networks structures have been proposed, of which convolutional neural network (CNN) is one of the most attractive models. CNN is an older deep neural network architecture \cite{cnn00}, and has enjoyed the great popularity as a efficient approach in character recognition \cite{cnn0}. Human speech, which is also a sequential signal, can be transformed into a feature map that we can take similarly as an image. Human speech signals are highly variable because of different speaking accents, different speaking styles and uncertain noises from the environment. For speech recognition, CNN has several advantages: (\expandafter{\romannumeral1}) human speech signal has local correlations in both time and frequency, CNN is well suited to exploit these correlations explicitly through a local connectivity. (\expandafter{\romannumeral2}) CNN has the ability to capture the frequency shift in human speech signal.

Some researchers proposed CNN can be used in speech recognition, and it has been proved that deep CNN has better performance over general feed-forward neural network or GMM-HMM in several speech recognition tasks \cite{cnn7,cnn8,cnn9,cnn10}. Most of previous application of CNNs in speech recognition only used fewer convolutional layers. For example, Abdel-Hamid~et~al. \cite{cnn1} used one convolutional layer, one pooling layer and a few full-connected layers. Amodei~et~al. \cite{rnn2} also used only three convolutional layers as the feature preprocessing layers. Some researcher has shown that CNN-based speech recognition which uses raw speech as input can be more robust \cite{cnn11}, and very deep CNNs has also been proved to show great performance in noisy speech recognition and Large Vocabulary Continuous Speech Recognition (LVCSR) tasks \cite{cnn2,cnn3,cnn9,cnn14}. Generally, very small filters with 3*3 kernels have recently been successfully applied in acoustic modeling in hybrid NN-HMM speech recognition system, and pooling layer has been proved to be replaced by full-connected convolutional layers and pooling has no highlights for LVCSR tasks \cite{cnn13}.

CNN has the ability to exploit the internal dependency of speech sequences while having the advantage of fewer parameters than RNN. This is to say, we can use less complex computational cost to achieve the same performance as RNN. With the datasets of human speech becoming larger, we firmly believe that overfitting would be less important but the convergence speed is what we always care about. Meanwhile, we find there are few work that discusses the convergence speed of different configurations in deep CNNs for ASR. Typical architecture of deep CNNs for ASR usually contains some fully-connected convolutional layers at the bottom, followed by several recurrent layers and fully-connected feedforward layers, but we find that, in practice, it's too slow to train this type of architecture for acoustic modeling. Recently, deep residual learning has been shown to obtain good convergence performance and compelling convergence in computer vision \cite{res1,res2}, which attributes to its identity mapping as the skip connection in the residual block. To explore how can we attain an architecture with faster convergence, we (\expandafter{\romannumeral1}) propose a novel deep recurrent convolutional networks and compare its performance with traditional deep convolutional recurrent networks, and (\expandafter{\romannumeral2}) apply deep residual learning in all of our experimental models. In detail, we compare the convergence speed of above three different architectures, besides, we evaluate three different architecutres combined with our proposed bidirectional statistical n-gram language model by PER. Our work has meaningful reference for who are also applying deep CNNs to attain a faster convergence speed in ASR.

This paper is organized as follows. First, we review the basics of commonly used models in ASR including recurrent neural network and convolutional neural network (Section \textrm{II}), then we explain how the end-to-end approach of connectionist temporal classification can be used to decode the output phonemes probability distribution (Section \textrm{III}). Besides, we propose a bidirectional statistical n-gram language model to rectify the output sequences of acoustic model in Section \textrm{IV}. Section \textrm{V} explains the setup and some training details in our experiments. Section \textrm{VI} presents the comparison of the convergence speed between traditional deep convolutional recurrent networks, our novel deep recurrent convolutional networks and those applied with deep residual learning. Finally, we show the evaluation results by minimum test PER of all experimental architectures in Section \textrm{VII}.
\section{Review of Neural Network}
\subsection{ELU Nonlinearity}
The most common functions applied to a neuron's output is ReLU \cite{other5} as a function of its input with $f(x)=max(0,x)$. ReLU behaves better than traditional non-linear functions such as sigmoid or tanh, since the mean value of ReLU's activations is not zero, some neurons in practice always become dead during backpropagation. Exponential linear unit (ELU) was introduced in \cite{other12}, in contrast to ReLU, ELU has negative values which pushes the mean of activations close to zero, that is to say, ELU can decrease the gap between the normal gradient and unit natural gradient and, therefore, speed up training. The expression of ELU nonlinearity is shown below in Equation~\ref{qg1}.
\begin{equation}
f(x)=\left\{\begin{array}{ll}x &\mbox{$if$ $x>0$}\\
      \alpha (exp(x)-1) &\mbox{$if$ $x\leqslant0$} \end{array}\right.
\label{qg1}
\end{equation}
Since our work is based on deep CNNs, we can take ELU as the non-linear function to faster our network's convergence. Fast learning has a great impact on performance of training large datasets.
\subsection{Recurrent Neural Network}
\begin{figure}[hbt]
\centerline{\includegraphics[width=0.8\linewidth]{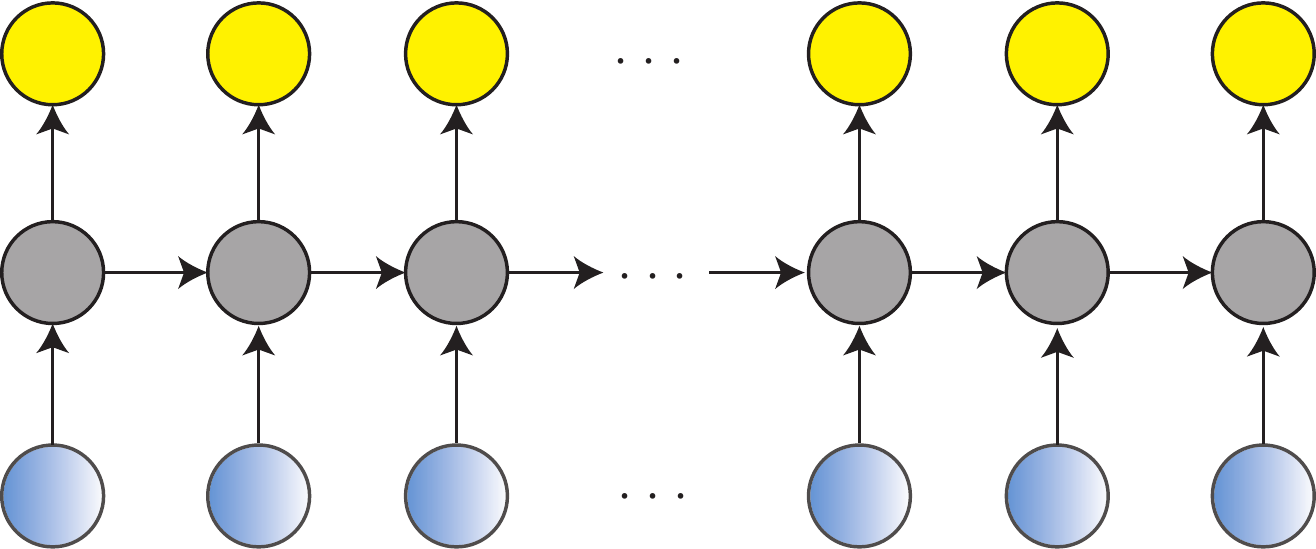}}
\caption{General architecture of a very simple RNN with a single hidden layer unfolded in time. Three layers shown above are input layer, recurrent hidden layer and output layer, respectively. The state of each hidden unit at t is decided by the state of hidden unit at t-1 and input at t.}
\label{rnn}
\end{figure}

General forward neural network can't depict the time dependency for sequence modeling problems well, such as automatic speech recognition. One type of special forward neural network is recurrent neural network(RNN). When RNN is folded out in time, it can be considered as a DNN with many sequential layers. In contrast to forward neural network, RNN is used to build time dependency of input features with internal memory units. RNN can receive input and produce output at every layer, the general architecture of RNN is shown in Figure~\ref{rnn}.  
As shown in Figure~\ref{rnn}, a very simple RNN is composed of an input layer, a hidden layer and an output layer. The recurrent hidden layer is designed to pass the forward information to backward time-steps. We can depict internal relationship of a general RNN in Equation~(\ref{qg_rnn}).
\begin{equation}
\textbf{O}^{t}=f(\textbf{y}^{t})=f(\textbf{W}^{hh}*\textbf{h}^{t-1}+\textbf{W}^{xh}*\textbf{x}^{t}+\textbf{b}^{t})
\label{qg_rnn}
\end{equation}
\noindent where $\textbf{W}^{hh}$ is the weight matrix between adjacent hidden units, $\textbf{h}^{t-1}$ is the hidden unit of previous time-step, $\textbf{W}^{xh}$ is the weight matrix between input layer and hidden layer, $\textbf{x}^{t}$ is the input at time t, and the bias $\textbf{b}^{t}$ is added and finally an activation function $f(\cdot)$, typically sigmoid, tanh, ReLU or ELU, will be applied to generate the output of the recurrent layer. If several recurrent layers are stacked, the output of previous layer becomes the input of next layer. 
\subsection{Convolutional Neural Network}
Compared to standard fully-connected neural networks and RNNs, CNNs which are proposed in \cite{cnn0} have much fewer free parameters so that they are easier to train. CNNs are pretty similar to ordinary neural networks, they are made of trainable weights and bias and they can also be stacked to a deep depth, which has been successfully applied in ImageNet competition \cite{ImageNet}.

\begin{figure}[hbt]
\centerline{\includegraphics[width=0.8\linewidth]{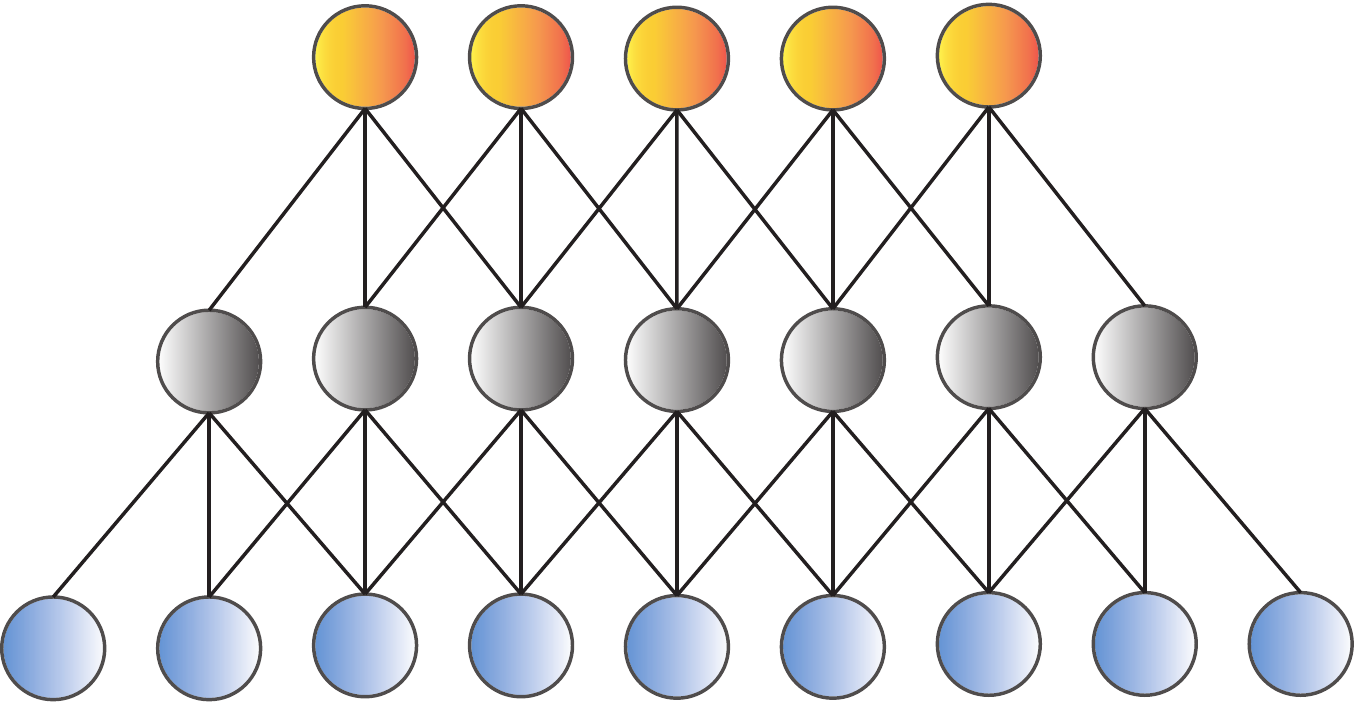}}
\caption{General architecture of 1-dimensional CNN. A typical CNN is composed of a convolutional layer and a pooling layer. We present a convolutional layer with filter size 1*3 and a pooling layer with pool size 1*3. We set the stride size to be 1 and no padding.}
\label{cnn}
\end{figure}

A typical architecture of simple CNN is composed of a convolutional layer and a pooling layer which is shown in Figure~\ref{cnn}. In most cases, a typical convolutional layer contains several feature maps, each of which is a kind of filter with shared parameters. These filters are spatial and extend through the full depth of input volume. Pooling layer is designed for dimensionality reduction and full-connected layer can output the probability distribution of all different classes. 

\begin{figure}[hbt]
\centerline{\includegraphics[width=0.8\linewidth]{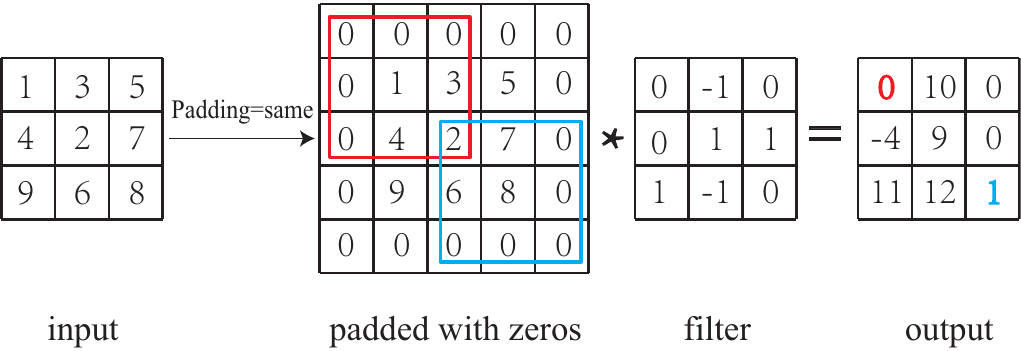}}
\caption{Example of convolution with padding in two dimensions. A 3*3 region is padded with zeros to a 5*5 region, then each 3*3 region is multiplied by the convolutional kernel to attain the output.}
\label{cnn_pad}
\end{figure}

In our experimented model, we replace the pooling layer with the full convolutional layer. Especially, we pad the input layer with zeros in both dimensions, since that the output layer can be the same size as the input layer if we set the stride to be 1. The details of padding is shown in Figure~\ref{cnn_pad}.
\section{Review of Connectionist Temporal Classification}
We also replace the traditional HMM decoder for output sequences. Labelling unsegmented sequential data is very common in speech recognition, and CTC is good at achieving this. The basic idea of CTC is to interpret the outputs of network as a probability distribution over all possible phenomes. Given this distribution, we can derive the objective function of sequence labeling. Since the objective function is differentiable, we can train it by backpropagation through time algorithm.

\begin{figure}[hbt]
\centerline{\includegraphics[width=0.8\linewidth]{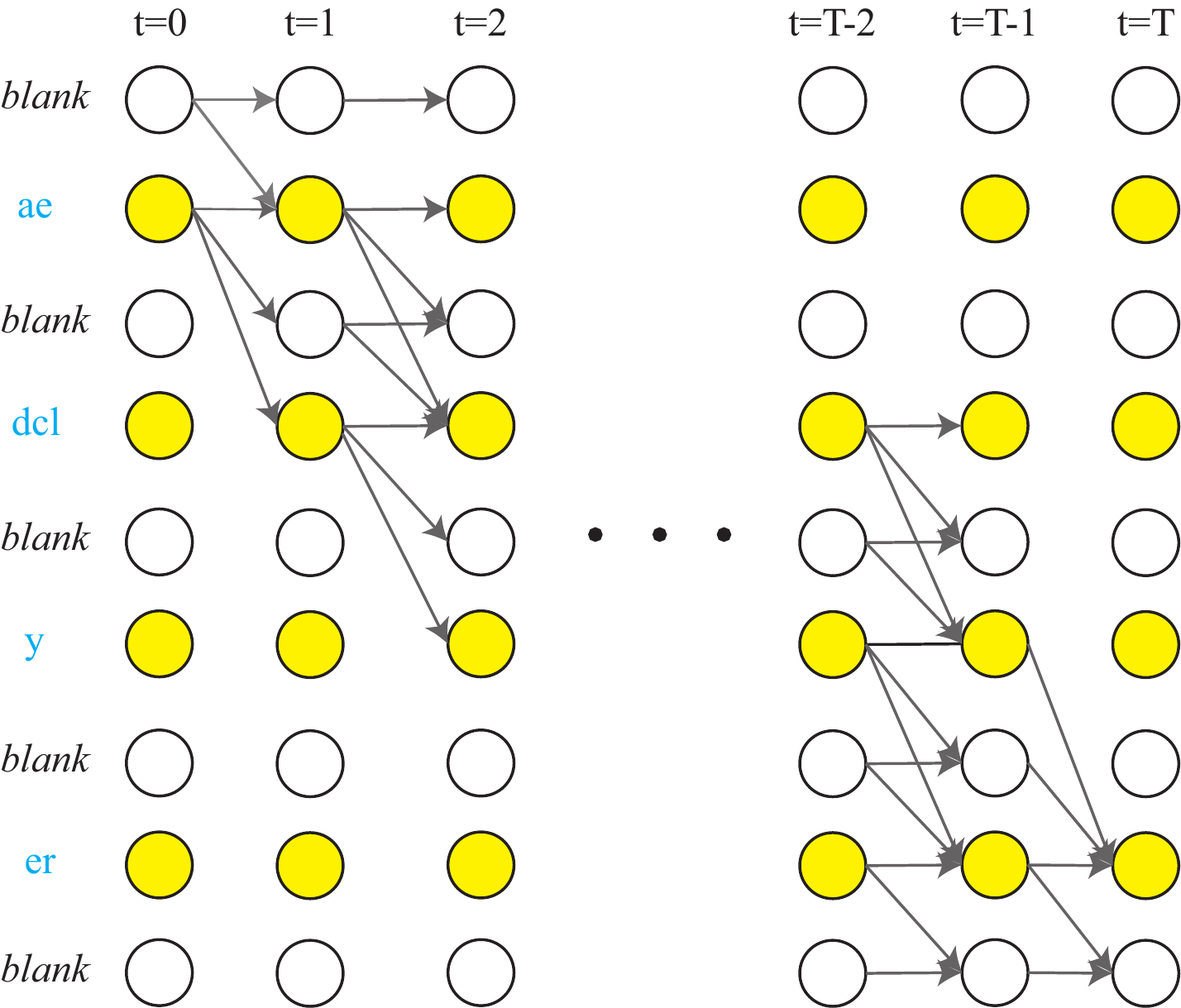}}
\caption{Illustration of the forward backward algorithm. White circle represents CTC blank, while yellow circle represents non-blank label. The direction of arrows represents what the forward variables are updated to, and backward variables are updated against them.}
\label{ctc}
\end{figure}

Using the probability distributions learned by deep CNNs, we would then use a CTC loss layer to finally output the phenome sequence. For a given input sequence, the goal of CTC is to minimize the average edit distance from the output sequence to actual sequence. Suppose $L$ is an alphabet of all output phonemes, in addition to $L$, the CTC network needs one more blank label, which is inserted between the adjacent two non-blank labels. The first $|L|$ activations stand for the probability of corresponding non-blank labels which we observed, and the probability of extra blank label is interpreted in the last activation unit. Given the output probability of CTC network, we can use beam search algorithm to compute the total probability of any sequence by considering its all paths as shown in Figure~\ref{ctc}. 

In detail, if we define $y_k^{t}$ as the probability of outputting label $k$ at time $t$, which defines a distribution over the set $L^{'T}$  of length T sequences over the alphabet $L^{'}=L\cap \{ blank \}$:
\begin{equation}
p(\pi\mid x)=\prod_{t=1}^{T}y_{\pi_{t}}^{t},\forall \pi \in L^{'T}
\label{qg2}
\end{equation}
In (\ref{qg2}), we take the elements of $L^{'T}$ as different paths, which is denoted as $\pi$. Given a labelling $l$ and a mapping function $B$ which removes all blanks and repeated labels, we can compute its probability by summing all its possible paths:
\begin{equation}
p(l\mid x)=\sum_{\pi \in B ^{-1}(l) }p (\pi \mid x)
\label{qg3}
\end{equation}
Therefore, the output sequence should be the most probable labelling, it's a decoding problem about how to find the output sequence. We require an efficient way of calculating the probabilities $p(l\mid x)$ of each labelling. Since from (\ref{qg3}) we may feel that it's very difficult to handle because there are many paths corresponding to a giving labelling. However, the problem can be converted to a dynamic programming problem, like what we use in HMMs. The key point of dynamic algorithm is to break down the sum over all paths into forward and backward variables.

To account for the blanks, CTC considers adding blanks between every pair of non-blank labels, including the beginning and the end. Therefore, the length of modified label $l'$ is $2|l|+1$. Since the state transition space for dynamic programming shouldn't be too large for computational efficiency, we assume that state transition only occurs between blank and non-blank label or two distinct non-blank labels. Besides, we allow all prefixes begins with a blank or the first symbol in $l$ and ends with a blank or the last symbol in $l$.
Thus, we can draw the expressions of dynamic program formulation from Figure~\ref{ctc}. Define the forward variable
\begin{equation}
\overline{\rm \alpha}_{t}(s)=\alpha_{t-1}(s)+\alpha_{t}(s-1)
\end{equation}
We can conclude that if the current label is blank or the current blank is the same as it was two steps ago, then the current forward variable is defined
\begin{equation}
\alpha_{t}(s)=\overline{\rm \alpha}_{t}(s)y_{l_{s}^{'}}^{t}
\end{equation}
Otherwise, the current forward variable is defined
\begin{equation}
\alpha_{t}(s)=(\overline{\rm \alpha}_{t}(s)+\alpha_{t-1}(s-2))y_{l_{s}^{'}}^{t}
\end{equation}
Since the last label must only be a blank or the last label in $l$, the final probability of a given label is calculated by forward variables as
\begin{equation}
p(l|x)=\alpha_{T}(|l|)+\alpha_{T}(|l|-1)
\end{equation}
The forward variables are defined previously, and the backward variables can be defined similarly. In practice, in order to avoid any underflows on any digital computer, we must normalize both the forward variables and backward variables. Finally, we can calculate the maximum likelihood error as
\begin{equation}
ln(p(l|x))=\sum_{t=1}^{T}ln(\sum_{s}\alpha_{t}(s))
\end{equation}
For forward and backward variables defined above, we can calculate the probability of any labels occurred at any time t given a labelling l as
\begin{equation}
\alpha_{t}(s)\beta_{t}(s)=y_{l_{s}}^{t}\sum_{\pi\in\beta^{-1}(l)}p(\pi|x)
\end{equation}
Since $p(l\mid x)$ can be obtained by summing over all s, differentiating this w.r.t $y_{k}^{t}$, we only consider paths that go through label k at time t.
\begin{equation}
\frac{p(l \mid x)}{y_{k}^{t}} = \frac{1}{{y_{k}^{t}}^{2}}\sum_{s \in pos(k,l)}\alpha_{t}(s)\beta_{t}(s)
\end{equation}
$pos(k,l)$ represents the set of positions where k occurs in l, so the objective function's gradient is
\begin{equation}
\frac{1}{y_{k}^{t}}(-ln(p(l \mid x))) = \frac{1}{p(l \mid x)} \frac{1}{{y_{k}^{t}}^{2} \sum_{s\in pos(k,l)}\alpha_{t}(s)\beta_{t}(s)}
\end{equation}
Our prediction of acoustic modeling is tied to CTC loss, and the loss takes the output of deep convolutional neural network as input. Using back propagation algorithm, we can update all parameters of acoustic model to reach a minimum of loss.
\section{Bidirectional N-gram Language Model}
\begin{figure*}[hbt]
\centerline{\includegraphics[width=0.9\linewidth]{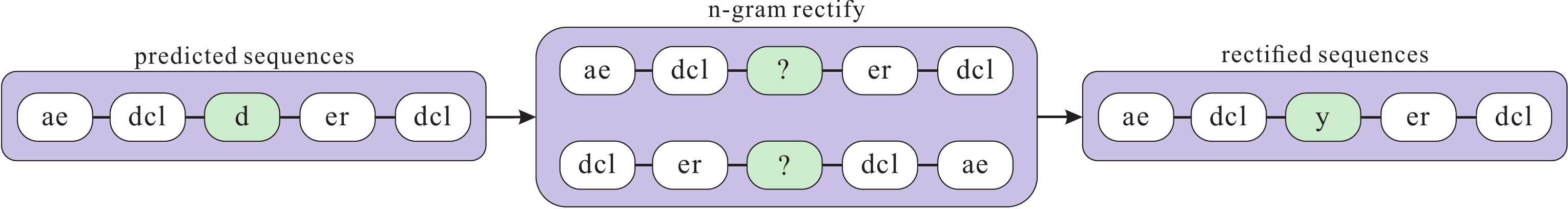}}
\caption{Example of the bidirectional statistical n-gram language model rectifying the output phoneme sequence.}
\label{n-gram}
\end{figure*}

Statistical language modeling and neural network have both been successfully used in speech recognition~\cite{Variable-ngram,Nplm,Rnnlm,Unngram}. Traditional n-gram model always makes an assumption that the probability of the current word depends only on the probability of the previous N-1 words, we propose a new bidirectional n-gram model which considers context probability of two sides. Computation of our bidirectional n-gram model is composed of two parts. First, We define forward n-gram going left to right in a sentence and we obtain the forward probability for each phoneme given previous phoneme phrase.
\begin{equation}
P_{f}(ph)=P(ph_{n}|ph_{n-1},...,ph_{n-(N-1)})
\end{equation}
\noindent Second, we reverse the whole training sentence to obtain the backward probability for each phoneme given future phoneme phrase.
\begin{equation}
P_{b}(ph)=P(ph_{n}|ph_{n+1},...,ph_{n+(N-1)})
\end{equation}
Our model takes as input bidirectional n-gram phoneme counts, thus it combines the bidirectional context information capacity and simple computation complexity. Besides, to make our model more robust, we perform bigram, trigram and four-gram including both forward probability and backward probability based on TIMIT corpus. We shows an example how our language model rectify mislabeled phonemes of acoustic model in Figure~\ref{n-gram}.
\section{Experimental Setup}
\subsection{Dataset}
We perform our speech recognition experiments on a public commonly used speech dataset: TIMIT. TIMIT is composed of 630 speakers with 6300 utterances of different sexes and dialects. Every audio clip is followed by phoneme transcriptions and sentence transcriptions, and we take the phoneme transcriptions as ground labels. The output is probability distribution of 62 labels including 61 non-blank phonemes and one blank label for CTC.
Since a typical machine learning dataset contains training set, validation set and test set, we design to split 6300 utterances of TIMIT into 5000 training utterances, 1000 validation utterances and 300 test utterances. For the generalization of our model, we randomly choose six different partitions of TIMIT and then select the best partition by cross validation. We build a simply baseline model of deep neural networks to evaluate the performance of six different partitions, and we choose the best partition whose test cost curve is going down both stably and fast.
\subsection{Feature Selection}
At the stage of pre-processing, each piece of speech is analyzed using a 25-ms Hamming window with a fixed overlap of 10-ms. Each feature vector of a frame are calculated by Fourier-transform-based filter-bank analysis, which includes 39 log energy coefficients distributed on 13 mel frequency cepstral coefficients (MFCCs), along with their first and second temporal derivatives. Besides, all feature data are normalized so that each vector dimension has a zero mean and unit variance. Since the feature matrix of each audio speech differs in time length, we pad each feature matrix with zeros to a max length.  
\subsection{Implementation}
We build our deep neural network models based on open-source library $Lasagne$, which is a lightweight library to build and train neural networks in Theano. We take our experiment on GPU Tesla K80 to speed up our training. For CTC part, we choose to use a C++ implementation of Baidu Research and we write some Python code to wrap it in Lasagne. Besides, all cross validation, feature generation, data loading, result analysis, visualization are implemented in Python code by ourselves.
\subsection{Training Details}
All our experimental models are trained by end-to-end stochastic gradient descent algorithm with a mini-batch of 32. In detail, since the Adam optimization method \cite{other13} is computationally efficient, requires little memory allocation and is well suitable for training deep learning models with large data, we adopt the Adam method with a learning rate of 0.00005 at the start of training. Instead of setting learning rate to be 0.001 as the paper said, we find that a learning rate of 0.00005 can make the divergence more stable in practice. As our experimental models are very deep, we would like to adopt some regulations to avoid overfitting. Recetnly, batch normalization \cite{other3} has shown a better regularization performance, however it would add extra parameters and needs heavy data augmentaion, which we would like to avoid. Instead, we add a dropout layer after the recurrent layers and after the first full-connected feedforward layer to prevent it from overfitting. To keep the sequential information for better acoustic modeling, we reserve the whole context information for end-to-end training instead of splitting each audio into frames of same length for frame-wise cross-entropy training. Besides, on the top layer of our experimental models, we set the activation function to be linear, for that CTC decoding has wrapped the softmax layer inside.
\subsection{Evaluation}
Since our proposed model is end-to-end and phoneme-level, we use phoneme error rate (PER) to evaluate the result. The PER is computed after the CTC network had decoded the whole output sentence into a sequence of phonemes. We then compute the Damerau-Levenshtein distance between our predicted sequence and the truth sequence to obtain the mistake we made. The average number of mistakes over the length of the phoneme sequence is just our PER. We evaluate our model on the test set finally.
\section{Architecture}
To explore the convergence properties of deep convolutional networks with recurrent networks, we have conducted experiments with different configurations. In this section, we'll discuss three typical experiments we have tried, which are novel deep recurrent convolutional networks, traditional deep convolutional recurrent networks and residual networks. We'll compare the convergence speed of them according to different configurations including number of layers, number of parameters, number of feature maps and applying residual learning.
\subsection{End-to-End Training}
Traditional training approach of acoustic modeling is based on frame-wise cross-entropy of predicted output and true label, which needs handy alignment between input frames and output labels. To avoid such high labor cost, our approach exploits the dynamic decoding method based on CTC which can perform supervised learning on sequence data and avoid alignment between input data and output label. We choose a commonly used dataset TIMIT to work on our acoustic model. TIMIT contains 6300 utterances of 630 speakers in 8 dialects, each audio contains phoneme transcription, word transcription and the whole sentence. We choose the phoneme transcription as labels for phoneme-level training.
\begin{figure}[hbt]
\centerline{\includegraphics[width=0.8\linewidth]{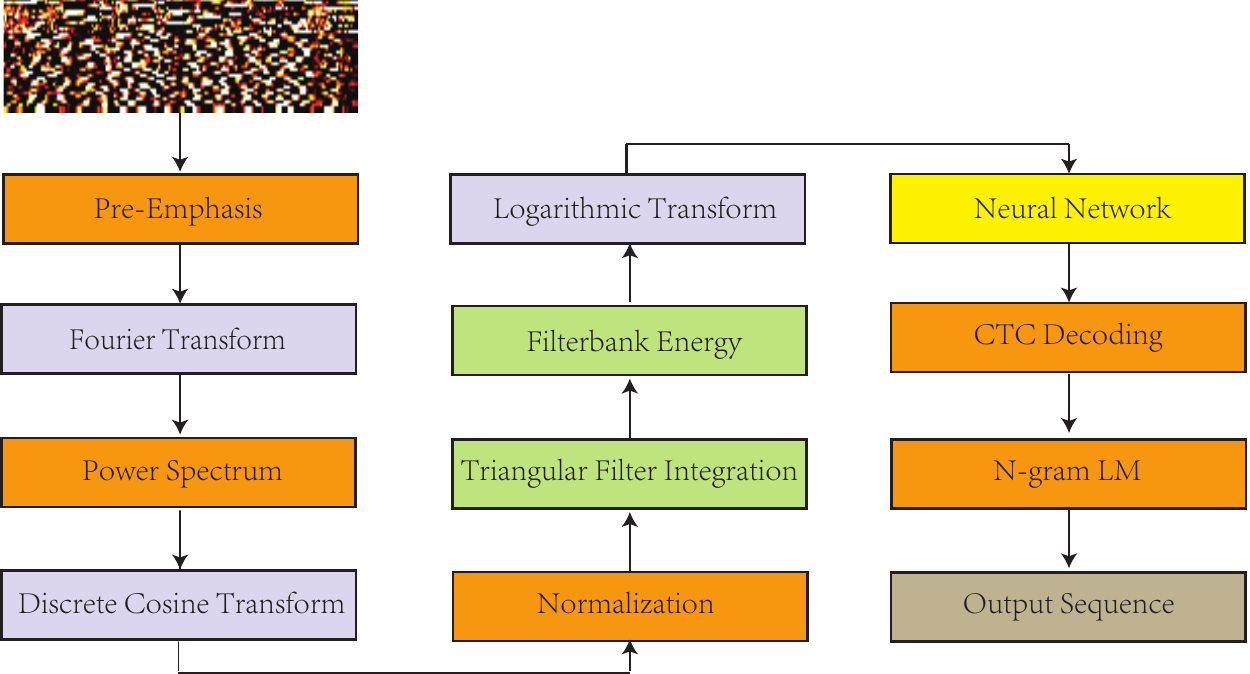}}
\caption{Overall process of our ASR system. We first calculate the mel frequency cepstral coefficients (MFCCs) of audio speech, and they are taken as input of experimental acoustic model. CTC is used for transcribe output probability distribution into phoneme sequence, and bidirectional statistical n-gram model is to rectify it.}
\label{mfcc}
\end{figure}
\subsection{Deep convolutional recurrent networks}
\subsubsection{Details of experimental models}
\begin{figure}[hbt]
\centerline{\includegraphics[width=0.9\linewidth]{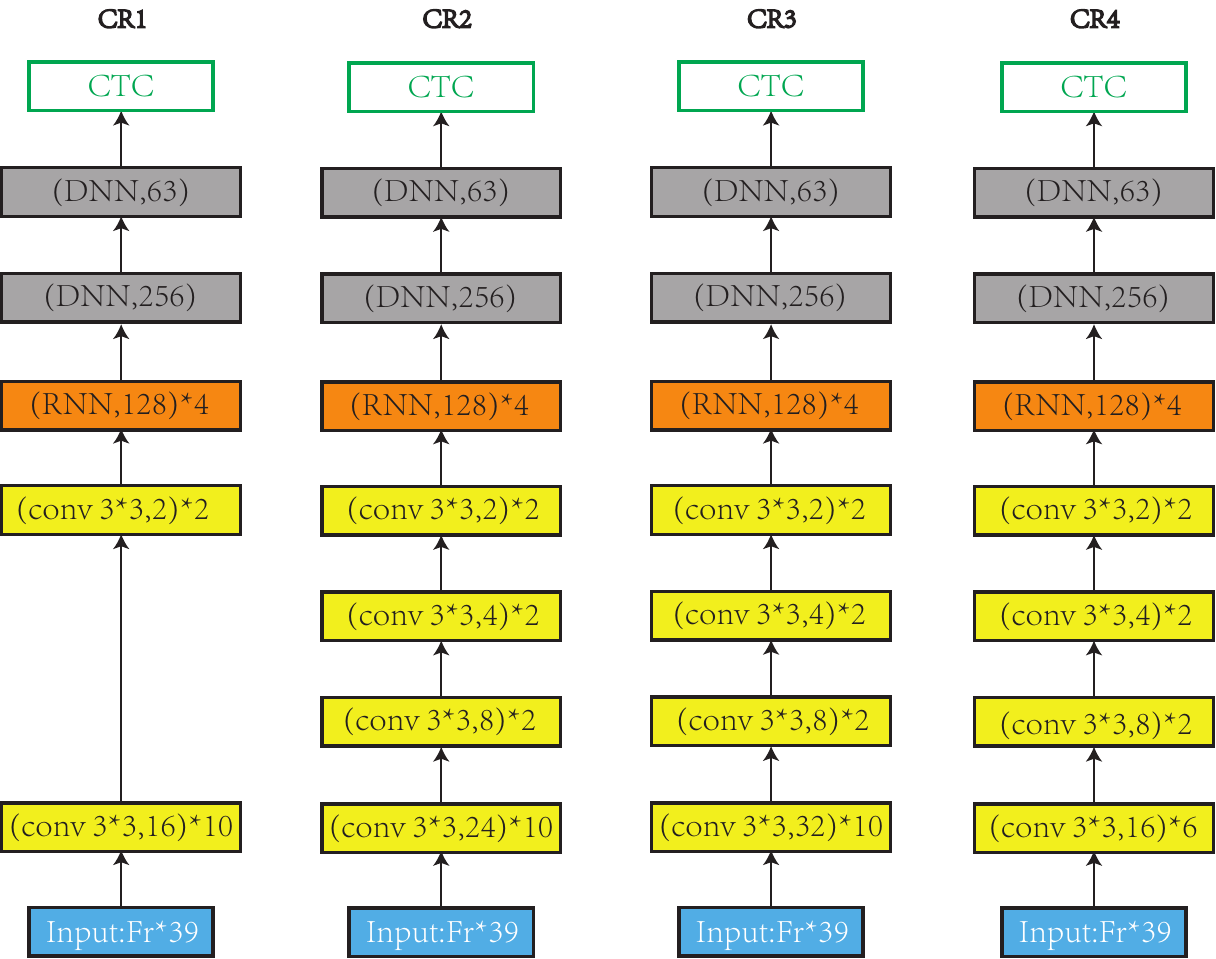}}
\caption{Architectures of four different deep convolutional recurrent models. We have conducted many experimental models, ``CR1'', ``CR2'', ``CR3'' and ``CR4'' are all traditional deep convolutional recurrent networks with better performance. }
\label{fig:all_models_2}
\end{figure}

Since many recent ASR systems use deep CNNs for acoustic modeling as part of ASR, especially, deep CNNs are used for feature preprocessing followed by RNN CTC decoding network. Conventional deep CNNs contain convolution and pooling, but we find that pooling can be replaced by convolution with fewer feature maps in practice. Inspired by this, we also build four deep convolutional recurrent networks, which are composed of deep convolutional layers, four recurrent layers and two full-connected feedforward layers. They are distinguished by different number of feature maps at convolutional layer. As shown in Figure~\ref{fig:all_models_2}, the two deep fully-connected convolutional recurrent networks ``CR1'' and ``CR2'' differ in number of feature maps in bottom convolutional layers. We set the number of feature maps to go down gradually, which means that the number of feature maps becomes narrower from bottom layers to up layers. Both ``CR3'' and ``CR4'' have narrower deep convolutional layers, but they have different bottom convolutional layers. The parameters of each model are shown in Table~\ref{tab:all_models_2}.  
\begin{table}[hbt]
\renewcommand{\arraystretch}{1.0}
\caption{Parameters of deep convolutional recurrent models}
\label{tab:all_models_2}
\centering
\begin{tabular}{c | c c c c}
\hline
\hline
Model & CR1 & CR2 & CR3 & CR4 \\
\hline
\#Params & 19k & 22k & 26k & 18k \\
\hline
\hline
\end{tabular}
\end{table}
\subsubsection{Comparison of cost curves}
\begin{figure}[hbt]
\centerline{\includegraphics[width=0.9\linewidth]{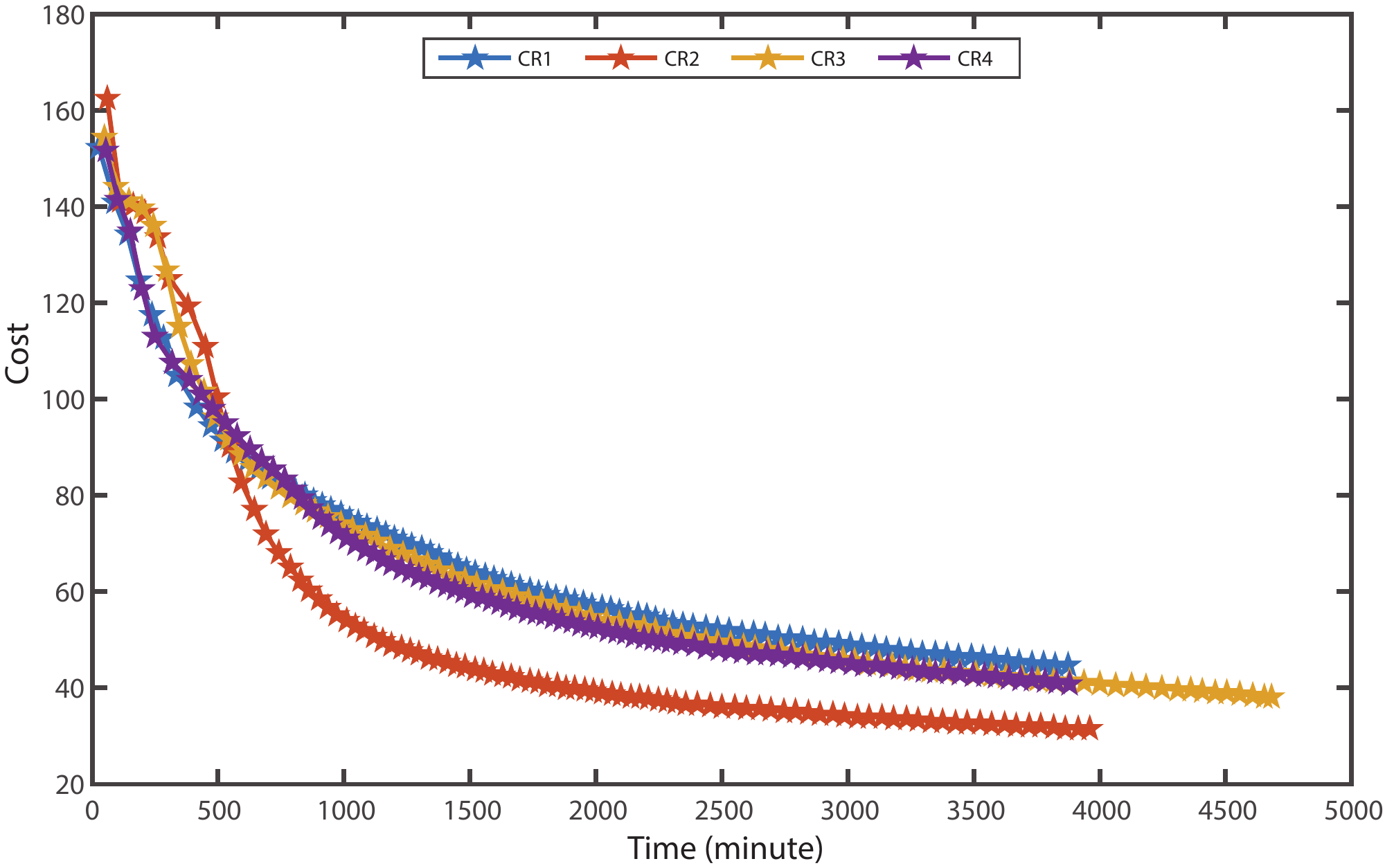}}
\caption{Cost curves of different convolutional recurrent models. We totally consider 91 epochs for each model. Since each model takes different time for an iteration, so two lines above have different length.}
\label{fig:convergence_CRNN}
\end{figure}

The cost curves of four convolutional recurrent models are shown in Figure~\ref{fig:convergence_CRNN}. Comparing to other three models, ``CR2'' converges fastest. Since ``CR1'' and ``CR2'' both have similar deep fully-connected convolutional layers, but differ in number of feature maps. We can find that ``CR2'' behaves much better than ``CR1''. ``CR3'' and ``CR4'' have narrower structures, they behave very similarly but much poorer than ``CR2''. So we can draw a conclusion that for convolutional recurrent networks, deep fully-connected convolutional layers can be of much help to convergence. However, narrower structure doesn't improve the performance slightly here.
\subsection{Deep recurrent convolutional networks}
Some previous ASR systems use pure stacked RNNs for acoustic modeling \cite{rnn1,rnn4}, and some recent ASR systems start to focus on taking some shallow CNNs as the stage of feature preprocessing in the bottom layers \cite{cnn8,rnn2}. We propose a new architecture for acoustic modeling which is composed of several recurrent layers followed by deep CNNs. Our new architecture has some highlights: First, we make use of RNNs to depict short time dependency of the input feature. Second, CNN is able to depict the local correlations of small field, but our deep stacked CNNs can see the whole context information. Third, as opposite to the traditionally used 6*6 or 3*4 filter and 1*3 pooling in speech recognition \cite{cnn9}, we use the small filters of 3*3 to build the full convolutional layers with no pooling layer. Filters of 3*3 is successfully used in computer vision, and we also find filters of 3*3 can effectively capture the high-level features along both time dimension and frequency dimension with little computational complexity. The stride of convolutional layer is set to be 1, and each convolutional layer is padded with zeros in both dimensions to keep the sizes of input and output feature maps unchanged. 
\begin{figure*}[hbt]
\centering
\includegraphics[width=0.9\linewidth]{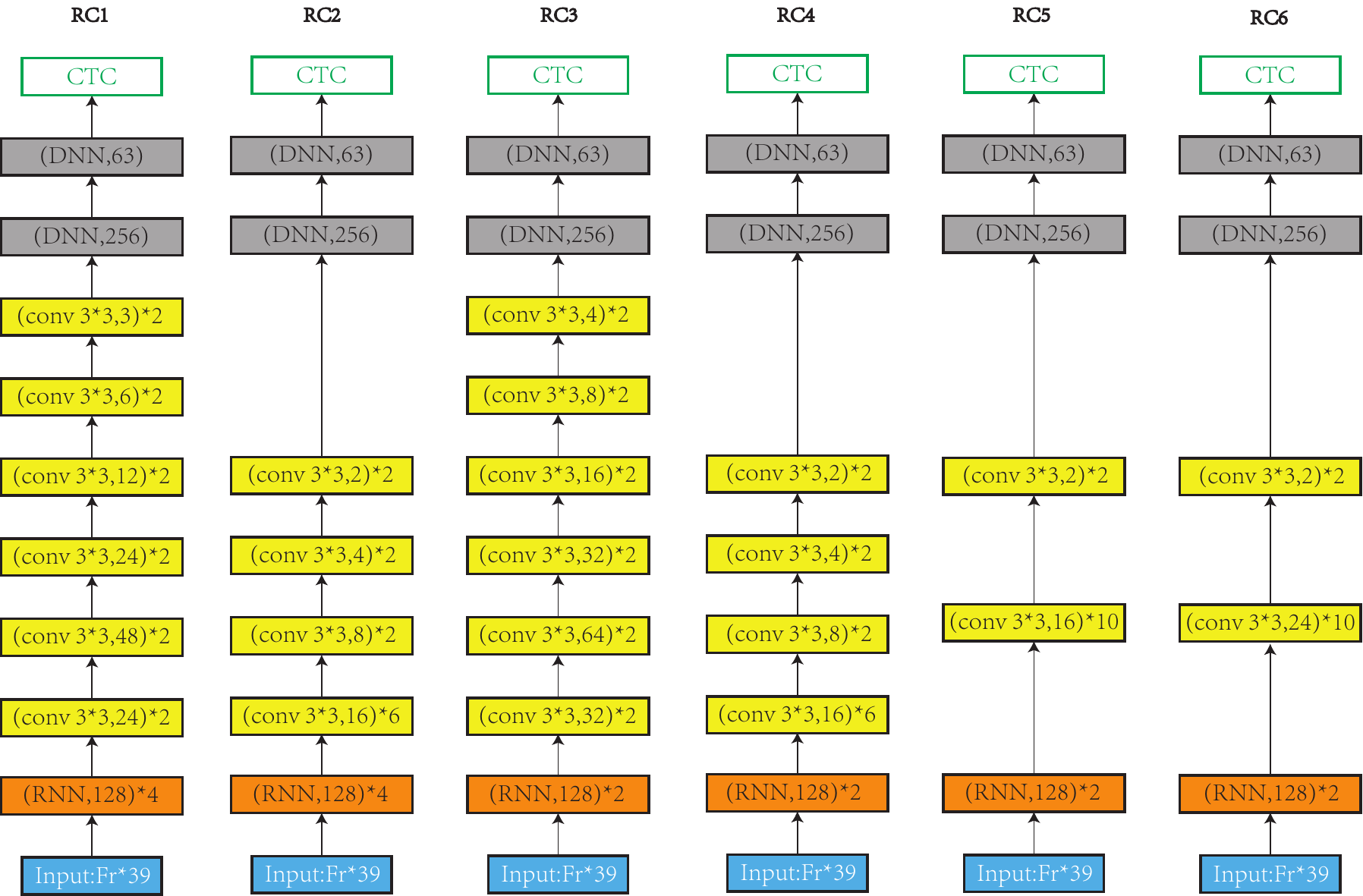}
\caption{Architectures of six different deep recurrent convolutional networks. We also conducted many experimental models of deep recurrent convolutional networks, and we show some six ones above.}
\label{fig:model1}
\end{figure*}
\subsubsection{Details of experimental networks}
We build six different networks which are combinations of RNNs and CNNs. The similarities of six networks are as follows. Continuous frames of 39 dimensional feature vectors are passed into the bottom recurrent layer as input of the network, and each recurrent layer has 128 neurons to store hidden information. Besides, at the top of each network, we add two full-connected feedforward layers, which have 256 hidden neurons and 62 output neurons, respectively. Since the CTC network has the softmax function, so we don't add any activation function on the top layer. Instead, we take the linear activation function as the output followed by CTC network. 

As shown in Figure~\ref{fig:model1}, ``RC1'' is a model with four recurrent layers at the bottom, followed by totally 12 convolutional layers, of which they are two convolutional layers with 24 feature maps, two convolutional layers with 48 feature maps, two convolutional layers with 24 feature maps, two convolutional layer with 12 feature maps, two convolutional layers with 6 feature maps and two convolutional layers with three feature maps. Compared with ``RC1'', we also build ``RC2'' with fewer parameters, and the difference between both is that ``RC2'' has 6 convolutional layers with 16 feature maps, 2 convolutional layers with 8 feature maps, 2 convolutional layers with 4 feature maps and 2 convolutional layers with 2 feature maps. Besides, we also build ``RC3'', which has the similar structure as ``RC1'' but with only 2 recurrent layers. Comparing with ``RC2'', we also build a similar network as ``RC4'', which has the same convolutional layers but with only 2 recurrent layers. Besides, we also build another two networks ``RC5'' and ``RC6'', which have more full-connected convolutional layers than previous four networks. The parameters of six networks are shown in Table~\ref{tab:params_rcnn}. 
\begin{table}[hbt]
\renewcommand{\arraystretch}{1.0}
\caption{Parameters of different recurrent convolutional models}
\label{tab:params_rcnn}
\centering
\begin{tabular}{c|c c c c c c}
\hline
\hline
Model&RC1&RC2&RC3&RC4&RC5&RC6\\
\hline
\#Params&29K&21K&23K&15K&15K&15K \\
\hline
\hline
\end{tabular}
\end{table}
\subsubsection{Comparison of cost curves}
\begin{figure}[hbt]
\centerline{\includegraphics[width=0.9\linewidth]{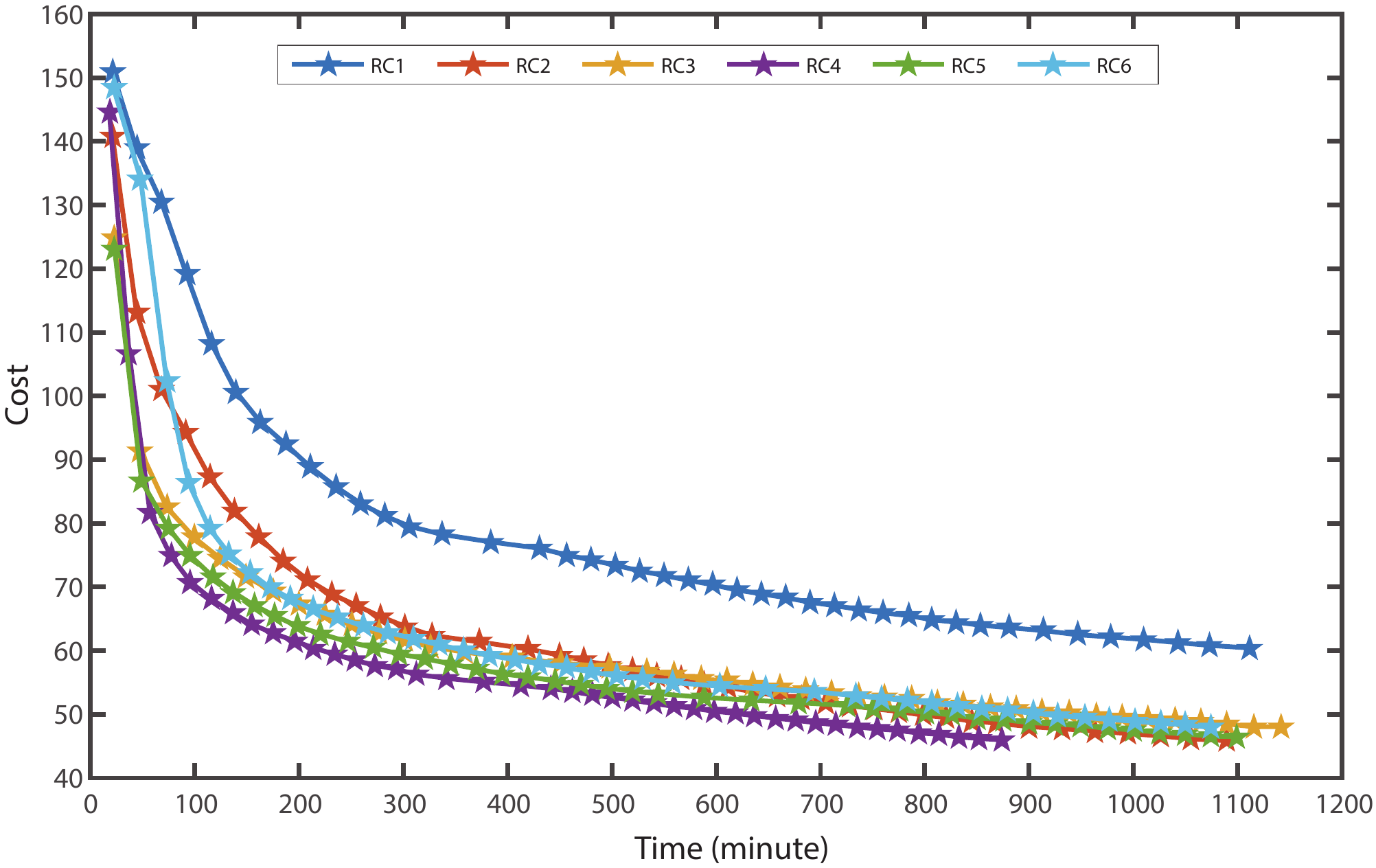}}
\caption{Cost curves of different recurrent convolutional architectures. We totally consider 42 epochs for every model. Since each model takes different time for an iteration, so two lines above have different length.}
\label{fig:cost1}
\end{figure}

To investigate the convergence of different networks in Figure~\ref{fig:model1}, we present the comparison of cost curves in Figure~\ref{fig:cost1}. The horizontal axis represents time, of which the unit is minute. The vertical axis represents training cost. We totally train each model for 42 epochs, and each marker denotes an epoch in Figure~\ref{fig:cost1}. Results show that ``RC1'' converges the slowest at both the beginning and the end, which is probably caused by too many parameters and too many convolutional layers of different feature maps. Differently, ``RC2'' performs much better than ``RC1'', since both finished 42 epochs at same time, but there is a great gap between their cost curves. Comparing ''RC1'' with ``RC2'', ``RC2'' has fewer parameters and more continuous full-connected convolutional layers. ``RC3'' has fewer recurrent layers than ``RC1'', and ``RC3'' also behaves much better than ``RC1''. Similarly, ``RC4'' has fewer recurrent layers than ``RC2'', and it behaves much better than ''RC2''. ``RC4'' finishes 42 epochs in only 900 minutes, but ''RC2'' takes around 1100 minutes. Since ``RC4'', ``RC5'' and ``RC6'' have the similar number of parameters, we find that ``RC4'' behaves best among three, which probably attributes to the different structure of ``RC4''. Based on ``RC5'', ``RC6'' has more convolutional feature maps, but we find this degrades the performance of ``RC5''.

Totally, our experiments show that narrower fully-connected convolutional layers can help improve model to converge, as ``RC2'' and ``RC4''. Besides, deep fully-connected convolutional layers with same number of feature maps have slightly inferior performance to narrower ones, as ``RC5'' and ``RC6''. For convergence, narrower recurrent convolutional layers behave better.  
\subsubsection{Comparison of ``RC2'' and ``CR2''}
\begin{figure}[hbt]
\centerline{\includegraphics[width=0.9\linewidth]{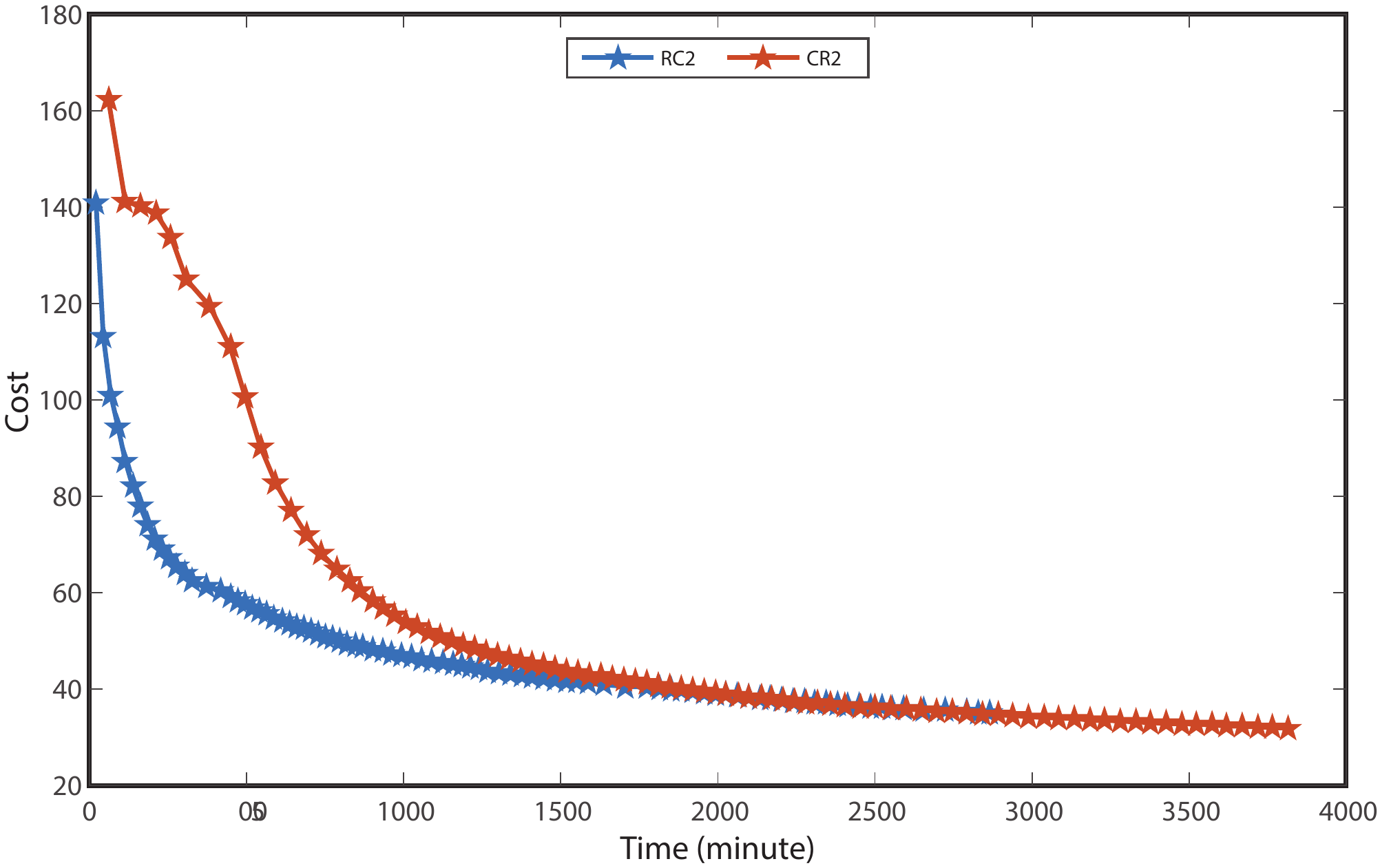}}
\caption{Cost curves of deep convolutional recurrent network ``CR2'' and deep recurrent convolutional ``RC2''. For both networks, we totally consider 88 epochs, respectively. Each pentagram marker denotes an epoch. Since each model takes different time for an iteration, so two lines above have different length.}
\label{fig:comparison_12}
\end{figure}

To investigate the difference of convergence between deep recurrent convolutional network and deep convolutional recurrent network, we present a comparison of convergence curves between both. Recurrent convolutional network is our novel proposed network for acoustic modeling, and convolutional recurrent network is the conventional one, they have the similar number of parameters. Since we have discussed some different models of them previously, we especially select two better models ``RC2'' and ``CR2'' to be shown in Figure~\ref{fig:comparison_12}.
As shown in Figure~\ref{fig:comparison_12}, we count 88 epochs for both ``CR2'' and ``RC2''. Observing the training cost curves along time, we can draw some conclusions. First, ``RC2'' finishes 88 epoches in less than 3000 minutes, while ``CR2'' finishes in nearly 4000 minutes, even though ``CR2'' has the same number of parameters as ``RC2'', but it converges one quarter faster than ``RC2''. Besides, in the first 2000 minutes, ``RC2'' behaves much better than ``CR2'', but in the last half of training, ``CR2'' catches up with ``RC2''. Therefore, our proposed deep recurrent convolutional network has the faster convergence performance at the beginning, and reach the same performance as conventional deep convolutional recurrent network later.
\subsection{Residual networks}
Generally, deeper convolutional neural networks can have larger capacity for feature representation, however, it has been shown that deeper convolutional neural networks are more difficult to train for their degradation problem, although there are some modern optimization methods. Recently, a new residual learning framework has been proposed to ease the training of very deep convolutional neural networks, and deep residual networks~\cite{res1} have been proved to improve convergence and higher accuracy in image classification with no more extra parameters. General deep residual networks (ResNets) are composed of many stacked ``Residual Blocks'', and each residual block can be expressed in following form:
\begin{equation}
y_{l} = h(x_{l})+\digamma(x_{l},W_{l})
\end{equation}
\begin{equation}
x_{l+1}=f(y_{l})
\end{equation}
where $x_{l}$ and $x_{l+1}$ are input and output of the $l$-th residual block, and $\digamma$ is a residual mapping function. In general, $h(x_{l})=x_{l}$ is an identity mapping and $f$ is an activation function, which we set to be $elu$~\cite{other12} here. With the presence of short-cut connection, residual networks can ease the degradation problem of deeper convolutional neural networks because the additional layers can only simply perform an identity mapping. Although ResNets with over 100 layers have shown great accuracy for several challenging image classification tasks on ImageNet competitions~\cite{ImageNet} and MS COCO competitions~\cite{COCO}, we also want to explore how the residual blocks behave in our experimental models. 
\begin{figure}[hbt]
\centerline{\includegraphics[width=0.9\linewidth]{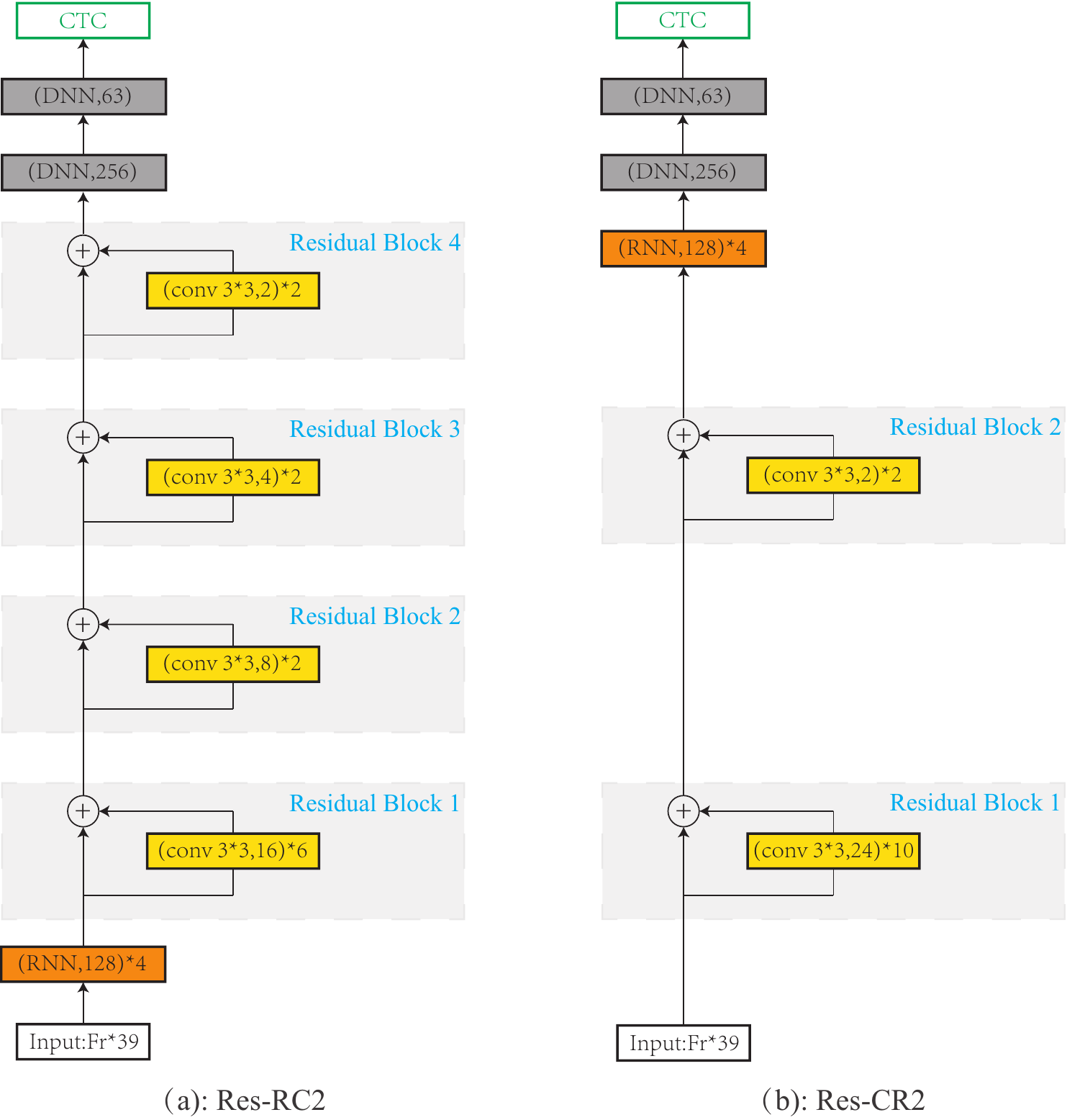}}
\caption{Architectures of both networks applied with deep residual learning framework. \textbf{(a)} We build four residual blocks based our novel network ``RC2'', each residual block contains layers with same number of feature maps. \textbf{(b)} We build two residual blocks based on traditional network ``CR2'', each residual block also contains layers with same number of feature maps.}
\label{fig:residual_models}
\end{figure}
To explore how ResNets behave in ASR, we propose a novel architecture which combines deep fully convolutional network with residual learning framework in ASR. We build two ResNets based on both ``CR2'' and ``RC2'', which are shown in Figure~\ref{fig:residual_models}. To avoid extra parameters, we build residual blocks only with layers of same dimension. ``CR2'' and ``RC2'' are two models with best performance among all discussed models above. For ``Res-RC2'', we build four residual blocks based on ``RC2'', and each residual block contains several layers with the same number of feature maps. In contrast, ``Res-CR2'' contains two residual blocks based on ``CR2''.
\begin{figure}[hbt]
\centerline{\includegraphics[width=0.9\linewidth]{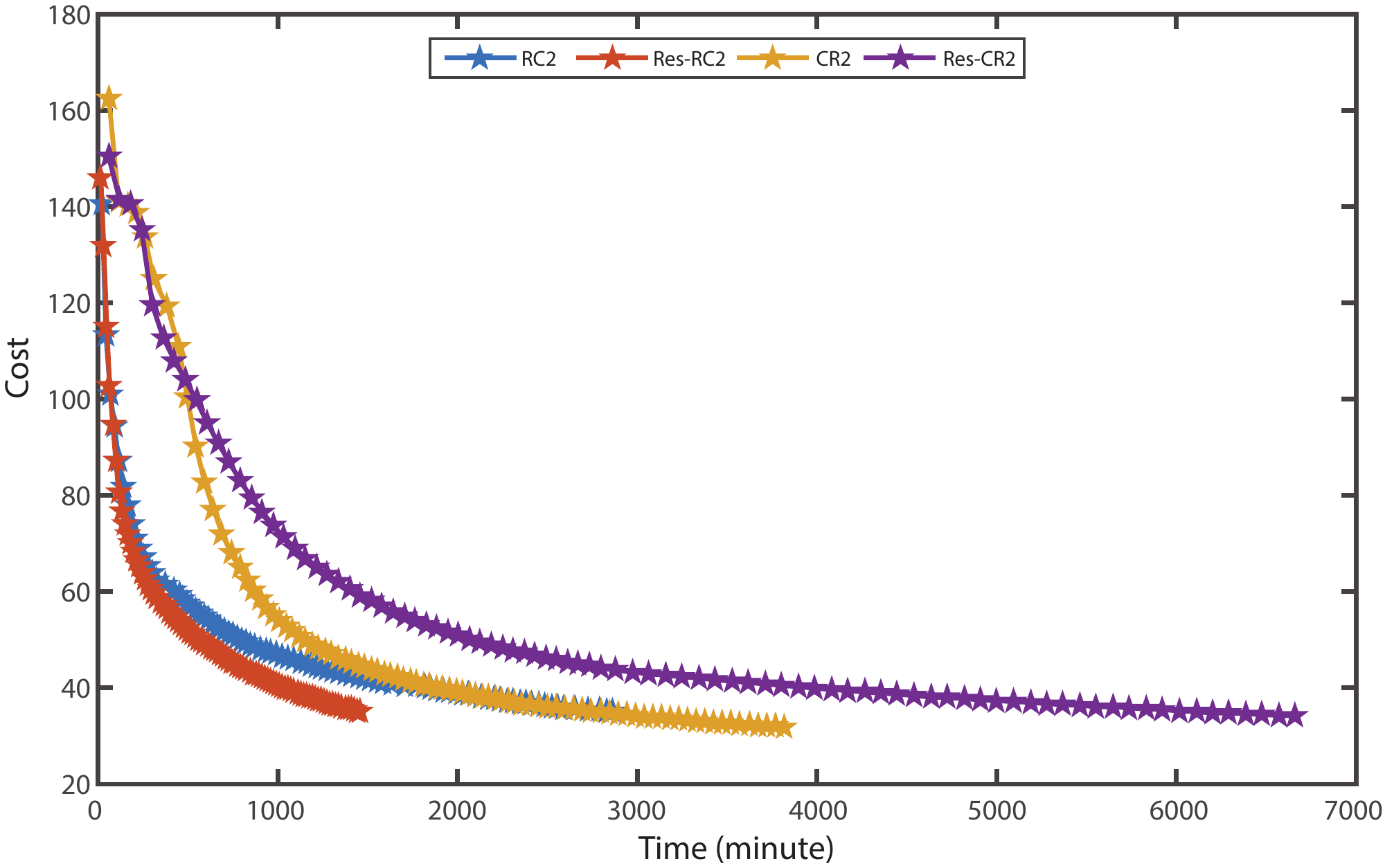}}
\caption{Cost curves of two plain networks and their residual versions. Each network is counted for 88 epochs.}
\label{fig:convergence_ResCRNN}
\end{figure}
Figure~\ref{fig:convergence_ResCRNN} gives the comparison of cost curves between two plain networks and their residual versions. ``Res-RC2'' shows the best performance, it finishes 88 epochs in only 1500 minutes, and its cost also goes down very quickly. Comparing ``Res-RC2'' with ``RC2'', we can draw a conclusion that ``Res-RC2'' converges twice as fast as ''RC2'', which mainly attributes to the four shortcuts connection of identity mapping in ``Res-RC2'', thus, the difficulty of training ``Res-CR2'' can be eased. However, compared to ``CR2'', ``Res-CR2'' with only two residual blocks converges much slower. Concretely, ``CR2'' finishes 88 epochs in less than 4000 minutes, while ``Res-CR2'' finished 88 epochs in nearly 7000 minutes. Although ``Res-CR2'' has two residual blocks based on ``CR2'', but the degradation problem has been exposed out of expectation. For this degradation problem, we propose two possible reasons. (\expandafter{\romannumeral1}) The bottom residual block of ``Res-CR2'' has ten convolutional layers, we think it's so deep that the power of deep residual learning may be restricted. (\expandafter{\romannumeral2}) ``Res-CR2'' has two recurrent layers on top of convolutional layers, so the convergence may be mainly influenced by the top recurrent layers and residual blocks make no difference here. 
\section{Evaluation}
TIMIT is a small 16 kHz speech corpus with 6300 utterances, from which the validation set of 300 utterances and the test set of 300 utterances are derived. We use 62 phonemes as output labels, including 61 non-blank labels and one blank label for CTC. After acoustic modeling, a probability distribution of 63 labels will be decoded by the CTC network into final phoneme sequences. Then, our proposed bidirectional hybrid n-gram language model over phonemes, estimated from the whole training set, is used to rectify the final sequence. 
\begin{table}[hbt]
\renewcommand{\arraystretch}{1.2}
\renewcommand{\multirowsetup}{\centering}  
\caption{Phoneme error rate of different architectures}
\label{tab:per_evaluation}
\centering
\begin{tabular}{ c | c | c | c }
\hline
\hline
Type & Model Structure & validation PER & test PER \\
\hline
\multirow{4}{2cm}{Deep convolutional recurrent networks} 
& CR1 & 21.56\% & 20.08\% \\
\cline{2-4}
& CR2 & \textbf{20.59\%} & \textbf{18.73\%} \\
\cline{2-4}
& CR3 & 23.89\% & 22.32\% \\
\cline{2-4}
& CR4 & 20.56\% & 19.31\% \\
\hline
\multirow{6}{2cm}{Deep recurrent convolutional networks} 
& RC1 & 32.17\% & 30.91\% \\
\cline{2-4}
& RC2 & \textbf{21.36\%} & \textbf{20.71\%} \\
\cline{2-4}
& RC3 & 25.66\% & 24.34\% \\
\cline{2-4}
& RC4 & 27.44\% & 25.54\% \\
\cline{2-4}
& RC5 & 25.76\% & 24.11\% \\
\cline{2-4}
& RC6 & 26.60\% & 24.87\% \\
\hline
\multirow{2}{2cm}{Deep Residual networks} 
& Res-RC2 & \textbf{18.77\%} & \textbf{17.33\%} \\
\cline{2-4}
& Res-CR2 & 20.13\% & 18.90\% \\
\cline{2-4}
\hline
\hline
\end{tabular}
\end{table}
Since our acoustic models are phoneme-level, we evaluate them by PER on the test set. Each experiment of an architecture has been conducted several times, we present the minimum validation PER and minimum test PER for every architecture in Table~\ref{tab:per_evaluation}. According to evaluation result, our novel deep recurrent convolutional network ``RC2'' obtains a PER of 20.71\%, with accuracy competitive to traditional deep belief network acoustic model. Although, deep convolutional recurrent network ``CR2'' achieves a test PER of 18.73\%, which is a 2\% relative improvement over the novel ``RC2'', however, when we apply the residual learning framework to them, we find that ``Res-RC2'' obtains a test PER of 17.33\%, which is an obvious improvement over ``RC2''. Furthermore, ``Res-CR2'' attains a PER of 18.90\% with slightly improvement over ``CR2'', which probably is caused by the heavy residual block as we have discussed previously. 
\section{Conclusions}
We propose a new architecture for sequence modeling in this work, and our novel deep recurrent convolutional network can have appealing performance on speech recognition task. In detail, we present an experimental comparison between two different acoustic models in speech recognition including traditional deep convolutional recurrent networks and our novel deep recurrent convolutional networks. Besides, we apply deep residual learning in both acoustic models. Traditional application of deep CNNs are used for the stage of feature preprocessing, followed by recurrent layers and CTC decoding layer. In contrast, it takes too much time to converge in practice. Our proposed deep recurrent convolutional network takes recurrent networks as feature preprocessing, and deep convolutional layers are designed to depict high-level feature representation. Experiments show that, compared to traditional deep convolutional recurrent networks, our novel deep recurrent convolutional network can converge in less time in the first half period and also attains a comparable PER. Besides, we try to apply deep residual learning in our acoustic models. We build some residual blocks through a shortcut connection as identity mapping for each model, and experiments show that our proposed novel deep recurrent convolutional networks can benefit a lot from these residual blocks both in accuracy and convergence. However, heavy residual blocks seem to have some negative impacts on the traditional deep convolutional recurrent networks in terms of convergence speed. Finally, we present a detailed analysis about their performance according to different training cost curves, and our proposed ``Res-RC2'' attains the best PER of 17.33\%. Our experiments verify that the novel deep recurrent convolutional networks can take place of traditional deep convolutional recurrent networks in ASR with less training time. In particular, deep residual learning can also be applied in the novel deep recurrent convolutional neural networks to make great improvement in both convergence speed and recognition accuracy. 
\section*{Acknowledgment}
The author would like to thank Chengyou Xie and Qionghaofeng Wu for helpful discussions on automatic speech recognition.

\end{document}